\pdfoutput=1

\documentclass[11pt]{article}

\usepackage{acl}

\usepackage{times}
\usepackage{latexsym}

\usepackage{graphicx}
\usepackage{arydshln}
\usepackage{amsmath}
\usepackage{bm}
\usepackage{booktabs}
\usepackage{pifont}
\usepackage{svg}
\usepackage{amssymb}
\usepackage{adjustbox}

\usepackage{tablefootnote}

\usepackage[T1]{fontenc}

\usepackage[utf8]{inputenc}

\usepackage{microtype}

\title{Encoding Sentence Position in Context-Aware Neural Machine Translation with Concatenation}

\author{
Lorenzo Lupo$^1$ \ Marco Dinarelli$^1$ \ Laurent Besacier$^{2}$ \\
$^1$Université Grenoble Alpes, France \\
$^2$Naver Labs Europe, France \\
\texttt{lorenzo.lupo@univ-grenoble-alpes.fr}\\
\texttt{marco.dinarelli@univ-grenoble-alpes.fr}\\
\texttt{laurent.besacier@naverlabs.com}
}

\begin{document}
\maketitle

\begin{abstract}
Context-aware translation can be achieved by processing a concatenation of consecutive sentences with the standard Transformer architecture.
This paper investigates the intuitive idea of providing the model with explicit information about the position of the sentences contained in the concatenation window.
We compare various methods to encode sentence positions into token representations, including novel methods. Our results show that the Transformer benefits from certain sentence position encoding methods on En$\rightarrow$Ru, if trained with a context-discounted loss \cite{lupo_focused_2022}. However, the same benefits are not observed on En$\rightarrow$De. Further empirical efforts are necessary to define the conditions under which the proposed approach is beneficial.
\end{abstract}

\section{Introduction}\label{se:intro}

Current neural machine translation (NMT) systems have reached human-like quality in translating standalone sentences, but there is still room for improvement when it comes to translating entire documents \cite{laubli_has_2018,castilho_context_2020}.
Researchers have attempted to close this gap by developing various context-aware NMT (CANMT) approaches, where \textit{context} refers to the sentences preceding or following the \textit{current} sentence to be translated.
A common approach to CANMT is sentence concatenation \cite{tiedemann_neural_2017, agrawal_contextual_2018, junczys-dowmunt_microsoft_2019}. The current sentence and its context are concatenated into a unique sequence that is fed to the standard Transformer architecture \cite{vaswani_attention_2017}.
Despite its simplicity, the concatenation approach has been shown to achieve competitive or superior performance to more sophisticated, multi-encoding systems~\cite{lopes_document-level_2020,lupo_divide_2022}.
However, learning with long concatenation sequences has been proven challenging for the Transformer architecture, because the self-attention can be "distracted" by long context \cite{zhang_long-short_2020,bao_g-transformer_2021}.

Recently, \citet{lupo_focused_2022} introduced the \textit{segment-shifted position embeddings} as a way to help concatenation approaches discerning the sentences concatenated in the processed sequence and improve attention's local focus.
Explicitly telling the model which tokens belong to each sentence is not a new idea, but an intuitive one that was already tested successfully in other tasks and approaches \cite{devlin_bert_2019,voita_context-aware_2018, zheng_towards_2020}.
We believe that encoding into token representations explicit information about the position of the sentences in the concatenation sequence can improve translation quality.
The temporal structure of the document constitutes essential information for its understanding and for the correct disambiguation of inter-sentential discourse phenomena.
This work investigates this intuitive idea by comparing various approaches to encoding sentence position in concatenation approaches.

Our contributions are the following:
(i) we compare segment-shifted position embeddings with three kinds of segment embeddings, evaluating their impact on the performance of the concatenation approach;
(ii) we propose and evaluate making sentence position encodings persistent over layers, adding them to the input of every layer in addition to the first;
(iii) we propose and evaluate fusing position embeddings and segment embeddings into a single vector where token and sentence positions are encoded in two orthogonal sets of dimensions, allowing a clearer distinction between them, along with memory savings.

To the best of our knowledge, this is the first comparative study on the employment of sentence position encodings for CANMT.
The sentence position encoding variants proposed are not found to improve the performance of the concatenation approach except for one specific setting where a context-discounted training loss is employed \cite{lupo_focused_2022}. More empirical studies are needed to clearly define the conditions under which the proposed approaches are beneficial to CANMT with concatenation.
Nonetheless, we find it useful to share these preliminary results with the scientific community. In fact, the proposed approaches are intuitive and easy to implement, hence something that many practitioners would presumably try. We hope that our findings can guide future research on sentence position encodings, by avoiding redundant experiments on failing settings. 
\section{Proposed approach}\label{se:approaches}

A common method for training a concatenation model and translating is by sliding windows \cite{tiedemann_neural_2017}.
The sliding concatenation approach sKtoK translates a window $\bm{x}_K^j =\bm{x}^{j-K+1}\bm{x}^{j-K+2}\cdot\cdot\cdot\bm{x}^{j-1}\bm{x}^j$, of $K$ consecutive sentences belonging to the source document, including the current ($j$th) sentence and $K-1$ context sentences, into $\bm{y}_K^j$.
In this work we only consider past context, although future context can also be present in the concatenation window.
At training time, the standard NMT loss is calculated over the whole output $\bm{y}_K^j$.
At inference time, only the translation $\bm{y}^j$ of the current sentence is kept, while the context translation is discarded.
Then, the window is slid by one sentence forward to repeat the process for the $(j+1)$th sentence and its context.

\subsection{Sentence position encodings}\label{se:approaches:se}

To improve the discernability of the sentences concatenated in the window, we propose to equip the sKtoK approach with sentence position encodings.
In particular, we experiment with segment-shifted position embeddings and three segment embedding methods. \textbf{Segment-shifted position embeddings} \cite{lupo_focused_2022} consist in a slight modification of the Transformer's token position scheme, where the original token positions are shifted by a constant factor every time a new sentence is encountered in the concatenation window.
The resulting positions are encoded with sinusoidal embeddings as for \citet{vaswani_attention_2017}.

We also experiment with \textbf{one-hot}, \textbf{sinusoidal}, and \textbf{learned segment embeddings}, like BERT's segment embeddings \cite{devlin_bert_2019}.
Segment embeddings encode the position $k$ of each sentence within the window of $K$ concatenated sentences into a vector of size $d$.
We attribute sentence positions $k=1,2,..., K$ starting from right to left.
The underlying rationale is always to attribute the position $k=1$ to the current sentence, no matter how many sentences are concatenated as context.
%
%
%
%
The simplest strategy to integrate segment embeddings (SE) with position embeddings (PE) and token embeddings (TE) is by adding them \cite{devlin_bert_2019}.
This operation requires that all three embeddings have same dimensionality $d_{model}$:
\begin{figure}[h]
\includegraphics[trim={2.65cm 0 0 0},clip,width=0.48\textwidth]{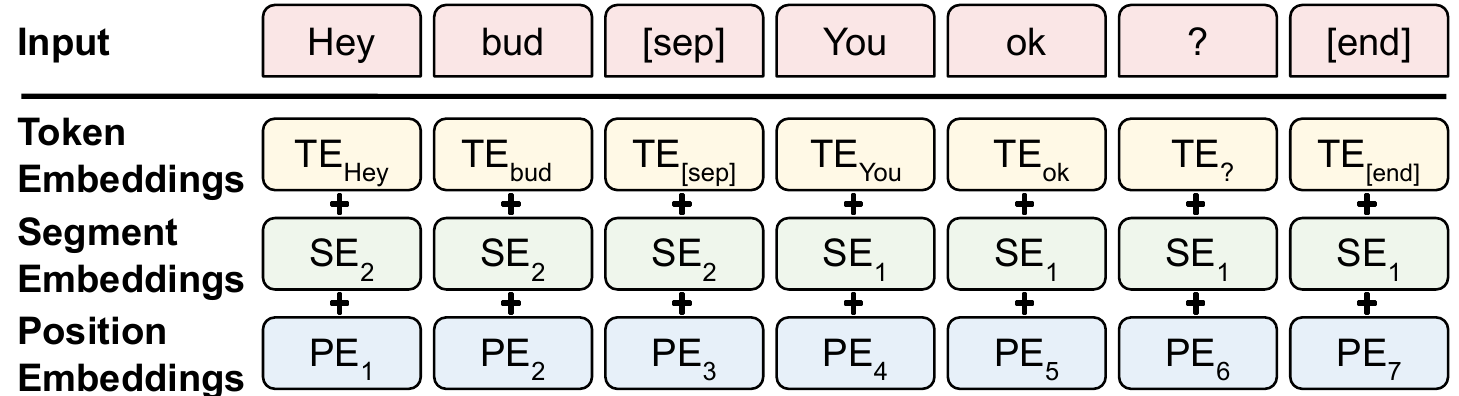}
\label{fig:se:pca}
\vspace{-6mm}
\end{figure}

\subsection{Persistent encodings}\label{se:approaches:pers}

We propose to make sentence position encodings persistent across Transformer's blocks, as \citet{liu_learning_2020} did for position embeddings. In other words, we propose adding segment-shifted position embeddings or segment embeddings to each block's input instead of limiting to the first one.

\subsection{Position-segment embeddings (PSE)}\label{se:approaches:pse}


In the Transformer, position embeddings are sinusoidal.
Their sum with the learnable token embeddings is based on the premise that the model can still distinguish both signals after being added up.
This distinction is accomplished by learning token embeddings in a way that guarantees them to be distinguishable.
Adding non-learnable segment embedding to this sum, however, rises the question whether they can be distinguished from the sinusoidal position embeddings. In some cases, learning to distinguish these two sources of information after their sum might be impossible. For instance, if segment embeddings are sinusoidal too, their sum with sinusoidal position embeddings is not bijective.\footnote{Consider, for example, the equivalence between, $PE_t + SE_k$ and $PE_k + SE_t$.}

Instead, concatenating PE and SE would make them perfectly distinguishable because they would belong to orthogonal spaces.
Unfortunately, concatenating two $d_{model}$-dimensional embeddings would then oblige to project the resulting vector back to a $d_{model}$-dimensional space.
To avoid this expensive operation, we propose to reduce the dimensionality of PE and SE from $d_{PE} = d_{SE} = d_{model}$ to values that sum up to the model dimension, i.e., $d_{PE} + d_{SE} = d_{model}$. Thus, each PE-SE pair can be concatenated into a unique vector named \textit{position-segment embedding} (PSE): $PSE_{t,k} = [PE_t, SE_k]$, of size $d_{model}$.\\
Reducing the dimensionality of PE and SE can be made without loss of information up to a certain degree, as it can be shown with a Principal Component Analysis \cite{jolliffe_principal_2016} of the sinusoidal position embedding matrix (Figure~\ref{fig:se:pca}).

\begin{figure}
	\includegraphics[width=0.45\textwidth]{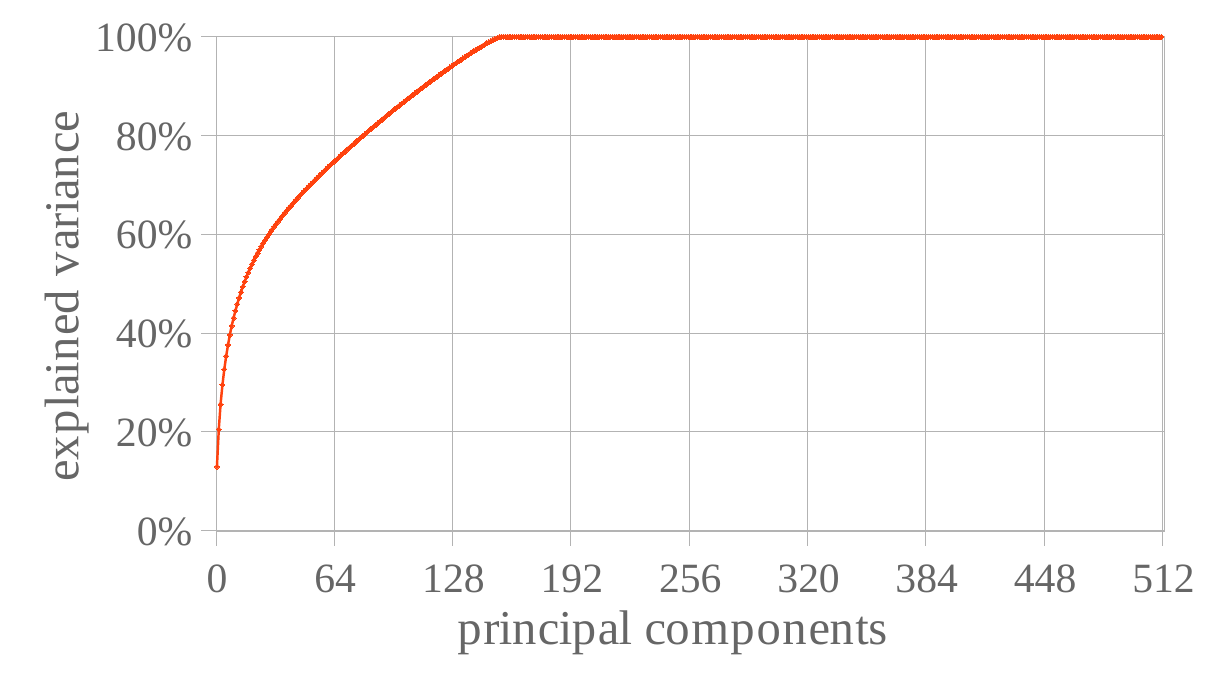}
	\caption{Cumulative ratio of the variance explained by the principal components of the of the sinusoidal position embedding matrix $PE\in\mathcal{R}^{1024\times 512}$, representing 1024 positions with 512 dimensions. Less than half of the principal components can explain the entirety of the variance represented in the sinusoidal embeddings. In other words, 1024 positions can be represented with the same resolution using less than half the dimensions.}
	\label{fig:se:pca}
\end{figure}


In the experimental section, we will empirically evaluate the impact of representing token and sentence positions with PSE, where the former are encoded with sinusoids and the latter with either one-hot, sinusoidal, or learned representations.


\section{Experiments}\label{se:exp}


We experiment with two models: \textbf{\textit{base}}, a context-agnostic \textit{Transformer-base} \cite{vaswani_attention_2017}, and \textbf{s4to4}, a context-sensitive concatenation approach with the same architecture as \textit{base}. s4to4 process sliding windows of 4 concatenated sentences in input and decodes the whole window into the target language.
We equip s4to4 with the sentence position encoding options presented in the previous Section, and we evaluate their impact on performance.
When experimenting with PSE, we allocate 4 dimensions to segment embeddings ($d_{SE} = 4$), which is enough to encode the position of each of the 4 sentences in the concatenation window, with both one-hot and sinusoidal encodings.
Since $d_{model} = 512$, this leaves $d_{PE}=508$ dimensions available to the sinusoidal representation of token positions.

The models are trained and evaluated on two language pairs covering different domains: En$\rightarrow$Ru movie subtitles prepared by \citet{voita_when_2019}, and  En$\rightarrow$De TED talk subtitles released by IWSLT17~(\citet{cettolo_wit3_2012}, see Table~\ref{tab:se:datastats} for statistics).
In addition to evaluating the average translation quality with BLEU\footnote{Moses' \textit{multi-bleu-detok}~\citep{koehn_moses_2007} for De, \textit{multi-bleu} for lowercased Ru as~\citet{voita_when_2019}.}, we employ two contrastive sets to evaluate the translation of context-dependent anaphoric pronouns.
For En$\rightarrow$Ru, we adopt \citet{voita_when_2019}'s set for the evaluation of inter-sentential deixis, lexical cohesion, verb-phrase ellipsis, and inflectional ellipsis.
For En$\rightarrow$De, we evaluate the models on the translation of context-dependent ambiguous pronouns with ContraPro~\citep{muller_large-scale_2018}, a large set of contrastive translations of inter-sentential pronominal anaphora.
Appendix~\ref{app:se:setup} includes more setup details. The implementation of our experiments is open-sourced on GitHub.\footnote{\href{https://github.com/lorelupo/focused-concat}{https://github.com/lorelupo/focused-concat}}


\subsection{Results}\label{se:exp:main}

\begin{table}[t]
\centering\begin{adjustbox}{width=0.925\linewidth}
	\begin{tabular}{lccccc}
		\toprule
		System & Enc. & Pers. & PSE & Voita & BLEU \\
		\midrule
		base &  &  &  & 46.64 & 31.98 \\
		s4to4 & \textbf{} & \textbf{} & \textbf{} & 72.02 & 32.45 \\
		\midrule
		s4to4 & shift &  &  & 71.28 & 32.27 \\
		s4to4 & shift & \checkmark &  & 71.80 & 31.93 \\
		\midrule
		s4to4 & 1hot &  &  & \textbf{72.52} & 32.61 \\
		s4to4 & 1hot & \checkmark &  & 71.44 & 32.42 \\
		s4to4 & 1hot &  & \checkmark & 71.24 & 32.33 \\
		s4to4 & 1hot & \checkmark & \checkmark & 71.16 & 32.41 \\
		\midrule
		s4to4 & sin &  &  & 71.92 & 32.39 \\
		s4to4 & sin & \checkmark &  & 71.20 & 32.38 \\
		s4to4 & sin &  & \checkmark & 71.26 & 32.56 \\
		s4to4 & sin & \checkmark & \checkmark & 71.68 & 32.38 \\
		\midrule
		s4to4 & lrn &  &  & 71.80 & 32.56 \\
		s4to4 & lrn & \checkmark &  & 71.40 & 32.50 \\
		s4to4 & lrn &  & \checkmark & 70.36 & 32.37 \\
		s4to4 & lrn & \checkmark & \checkmark & \textbf{73.20} & 32.38 \\
		\bottomrule
	\end{tabular}
\end{adjustbox}
\caption{En$\rightarrow$Ru models' accuracy on Voita's contrastive set and BLEU on the test set.
	s4to4 models are equipped with sentence position encodings (Enc.) of four kinds: segment-shifted position embeddings, one-hot segment embeddings, sinusoidal segment embeddings, or learned segment embeddings. Persistent encodings (Pers.) are added to the input of each Transformer's block. Alternatively to being added, segment embeddings can be concatenated with position embeddings (PSE). Values in bold are the best within their block of rows and outperform the baselines (base, s4to4).}\label{tab:se:ru}
\end{table}

First, we study the impact of sentence position encodings in the En$\rightarrow$Ru setting. In Table~\ref{tab:se:ru}, we compare models equipped with different combinations of encodings (Enc.) and integration methods: persistency (Pers.) and fusion with position encodings (PSE). 
We primarily focus on the contrastive evaluation of discourse translation since average translation quality metrics like BLEU have been repeatedly shown to be ill-equipped to detect improvements in CANMT \cite{hardmeier_discourse_2012}.
Indeed, BLEU displays negligible fluctuations throughout the whole table.
However, the performance on the contrastive sets is not encouraging either: most of the encoding variants degrade s4to4’sperformance.
The one-hot encoding helps, but only by a thin margin.
Making encoding persistent or concatenating them into PSE does not help either.
The only exception is s4to4+lrn+pers+PSE (last line), which gains more than two accuracy points over baseline.
However, this result is solely driven by the net improvement on deixis disambiguation (almost +5 points, see Table~\ref{tab:se:discdetails_ru}), while the performance is degraded on the other three discourse phenomena.
In conclusion, sentence position encodings do not seem to benefit the vanilla s4to4 approach.

\subsubsection{Training with context-discounted loss}\label{se:exp:main:cd}

\begin{table}[t]
\centering
\begin{adjustbox}{width=0.925\linewidth}
	\begin{tabular}{lccccc}
		\toprule
		System & Enc. & Pers. & PSE & Voita & BLEU \\
		\midrule
		base &  &  &  & 46.64 & 31.98 \\
		s4to4 &  &  &  & 72.02 & 32.45 \\
		s4to4+\texttt{CD} &  &  &  & 73.42 & 32.37 \\
		\midrule
		s4to4+\texttt{CD} & shift &  &  & 73.56 & 32.45 \\
		s4to4+\texttt{CD} & shift & \checkmark &  & \textbf{75.94} & 31.98 \\
		\midrule
		s4to4+\texttt{CD} & 1hot &  &  & 73.06 & 32.35 \\
		s4to4+\texttt{CD} & 1hot & \checkmark &  & 73.90 & 32.56 \\
		s4to4+\texttt{CD} & 1hot & \checkmark & \checkmark & \textbf{74.50} & 32.33 \\
		\midrule
		s4to4+\texttt{CD} & sin &  &  & 73.48 & 32.53 \\
		s4to4+\texttt{CD} & sin & \checkmark &  & 73.40 & 32.52 \\
		s4to4+\texttt{CD} & sin & \checkmark & \checkmark & \textbf{74.68} & 32.27 \\
		\midrule
		s4to4+\texttt{CD} & lrn &  &  & 73.68 & 32.45 \\
		s4to4+\texttt{CD} & lrn & \checkmark &  & \textbf{75.56} & 32.43 \\
		s4to4+\texttt{CD} & lrn & \checkmark & \checkmark & 74.48 & 32.35 \\
		\bottomrule
	\end{tabular}
\end{adjustbox}
\caption{En$\rightarrow$Ru context-discounted s4to4's accuracy on Voita's contrastive set and BLEU. Values in bold are the best within their block of rows and outperform the baselines (base, s4to4, s4to4+CD).}\label{tab:se:cd_ru}
\end{table}
Following \citet{lupo_focused_2022}, we hypothesize that sentence position encodings can be leveraged more effectively by training the concatenation approach with a context-discounted objective (see Appendix~\ref{app:se:cd} for details).
Indeed, the context-discounted objective function incentivizes distinguishing among different sentences.
Table~\ref{tab:se:cd_ru} displays the results of the s4to4+\texttt{CD} model equipped with the various combinations of encodings tested before, except the \textit{non-persistent} PSE.\footnote{Since preliminary experiments where not encouraging, we do not provide results for the non-persistent PSE combination in order to economize experiments.}
In this case, too, vanilla sentence encoding methods do not significantly help the s4to4+\texttt{CD} model.
However, making the encodings persistent boosts performance in the case of segment-shifted positions (+2.52 accuracy points over s4to4+\texttt{CD}) and learned embeddings (+2.14).
One-hot segment embeddings benefit only slightly (+0.48) from being persistent, while no improvement is measured in the case of sinusoidal segment embeddings.
As discussed in Section~\ref{se:approaches:pse}, this was expected since one-hot or sinusoidal segment embeddings might not be distinguishable from sinusoidal position embeddings once they are added together.
Instead, when one-hot and sinusoidal segment embeddings are concatenated to position embeddings into a unique PSE and made persistent, they boost s4to4+\texttt{CD} by +1.08 and +1.26 accuracy points, respectively.


\begin{table}[t]
\centering
\begin{adjustbox}{width=\linewidth}
	\begin{tabular}{lccccc}
    \toprule
    System & Enc. & Pers. & PSE & ContraPro & BLEU \\
    \midrule
    base &  &  &  & 43.57 & 29.63 \\
    s4to4 &  &  &  & 72.12 & 29.48 \\
    s4to4+\texttt{CD} & \textbf{} & \textbf{} & \textbf{} & 74.78 & 29.32 \\
    \midrule
    s4to4+\texttt{CD} & shift &  &  & 74.56 & 29.20 \\
    s4to4+\texttt{CD} & shift & \checkmark &  & 71.46 & 27.50 \\
    \midrule
    s4to4+\texttt{CD} & sin &  &  & 74.46 & 29.23 \\
    s4to4+\texttt{CD} & sin & \checkmark &  & 74.35 & 29.26 \\
    s4to4+\texttt{CD} & sin & \checkmark & \checkmark & 74.02 & 28.73 \\
    \midrule
    s4to4+\texttt{CD} & lrn &  &  & 72.49 & 28.35 \\
    s4to4+\texttt{CD} & lrn & \checkmark &  & 71.07 & 27.87 \\
    s4to4+\texttt{CD} & lrn & \checkmark & \checkmark & 71.89 & 28.63 \\
    \bottomrule
	\end{tabular}
\end{adjustbox}
\caption{Accuracy on ContraPro of models trained on En$\rightarrow$De IWSLT17, and BLEU on the test set.}\label{tab:se:cd_de}
\end{table}

With the aim of evaluating the generalizability of these results to another language pair and domain, we train the context-discounted approach on the En$\rightarrow$De IWSLT17 dataset and evaluate it on ContraPro \cite{muller_large-scale_2018}.\footnote{We don't experiment again with one-hot encodings since it was the less promising approach on the En$\rightarrow$Ru setting.}
Table~\ref{tab:se:cd_de} summarizes the results.
Unfortunately, the improvements achieved on En$\rightarrow$Ru do not transfer to this setting.
The s4to4+\texttt{CD} slightly benefits from segment-shifted position embeddings, but the other approaches degrade its performance.
We hypothesize that the model does not undergo sufficient training in this setting to reap the benefits of sentence position encodings.
In En$\rightarrow$De IWSLT17, the training data volume is smaller than in the En$\rightarrow$Ru setting by an order of magnitude: 0.2 million sentences versus 6 million (see Table~\ref{tab:se:datastats}).
Therefore, we extended the experiments on En$\rightarrow$De by training models on millions of sentences.
The details and results are presented in Appendix~\ref{app:se:nei} and Table~\ref{tab:se:nei}.
Unfortunately, even in this case, the En$\rightarrow$De s4to4+\textit{CD} does not benefit from the proposed sentence position encoding options.

\section{Benchmarking}\label{se:conclusions}
\begin{table}[]
	\centering
	\scalebox{0.95}{
	\begin{tabular}{lc}
		\toprule
		System\tablefootnote{\label{foot:sota}Whenever the cited works present and evaluate multiple systems, we compare to the best performing one. For the majority of these works, BLEU scores are not available for comparison on the same test set.} & Voita \\
		\midrule
		\citet{chen_breaking_2021} & 55.61 \\
		\citet{sun_rethinking_2022} & 58.13 \\
		\citet{zheng_towards_2020} & 63.30 \\
		\citet{kang_dynamic_2020} & 73.46 \\
		\citet{zhang_long-short_2020} & 75.61 \\
		s4to4 + shift$_{\textit{pers}}$ + \texttt{CD} & \textbf{75.94} \\
		\bottomrule
	\end{tabular}
}\caption{Benchmarking on En$\rightarrow$Ru (accuracy).}\label{tab:sota_ru}
\end{table}

\begin{table}[]
	\centering
	\scalebox{0.95}{
	\begin{tabular}{lc}
		\toprule
		System\textsuperscript{\ref{foot:sota}} & ContraPro \\
		\midrule
		\citet{maruf_selective_2019} & 45.04 \\
		\citet{voita_context-aware_2018}\tablefootnote{Reported in \citet{muller_large-scale_2018}.} & 49.04 \\
		\citet{stojanovski_improving_2019} & 57.64 \\
		\citet{muller_large-scale_2018} & 59.51 \\
		\citet{lupo_divide_2022} & 61.09 \\
        \citet{lopes_document-level_2020} & 70.8 \\
		\citet{majumder_baseline_2022} & 78.00 \\
		\citet{fernandes_measuring_2021} & 80.35 \\
		\citet{huo_diving_2020} & \textbf{82.60} \\
		s4to4 + \texttt{CD} & \textbf{82.54} \\
		\bottomrule
	\end{tabular}
}\caption{Benchmarking on En$\rightarrow$De (accuracy).}\label{tab:sota_de}
\end{table}

In Tables \ref{tab:sota_ru} and \ref{tab:sota_de}, we compare our best performing systems with other CANMT systems from the literature. For En$\rightarrow$Ru (Table \ref{tab:sota_ru}), we compare with works that adopted the same experimental conditions as ours. Our s4to4 concatenation approach trained with context discounting and persistent segment-shifted positions achieves the best accuracy on Voita's contrastive set. For En$\rightarrow$De (Table \ref{tab:sota_de}), we compare to the works adopting \citet{muller_large-scale_2018}'s contrastive set for evaluation, even if the training conditions are not comparable. Our s4to4+\texttt{CD} trained on the high resource setting (see Appendix~\ref{app:se:nei}) is second of the list, by a negligible margin. Notably, \citet{huo_diving_2020}'s system is also a concatenation approach, but trained on x10 parallel sentences with respect to our system. This comparison indicates that context discounting \cite{lupo_focused_2022} makes training efficient. 


\section{Conclusions}\label{se:conclusions}


Intending to improve concatenation approaches to context-aware NMT (CANMT), we investigated an intuitive idea: encoding into token representations the position of their sentence within the processed sequence.
Besides adopting existing encoding methods (segment-shifted position embeddings and segment embeddings), we proposed a novel approach to integrate token and sentence position embeddings in a unique vector called position-segment embedding (PSE). We also propose to make sentence position encodings persistent throughout the model's layers.

We compared these encoding approaches on the En$\rightarrow$Ru/De language pairs.
Consistent improvements were observed on En$\rightarrow$Ru when persistent sentence position encoding methods were used in conjunction with the context-discounted training objective proposed by \citet{lupo_focused_2022}. However, results on En$\rightarrow$De were negative.

Further research is needed to clearly define the conditions under which the proposed approaches are beneficial to CANMT with concatenation.
We encourage practitioners to test the most promising sentence-position encodings - \textbf{persistent segment-shifted positions} - should they want to get the most out of their CANMT systems, but only in conjunction with \textbf{context discounting}.

\section*{Acknowledgements}
We thank the anonymous reviewers for their insightful comments.
This work has been partially supported by the Multidisciplinary Institute in Artificial Intelligence MIAI@Grenoble Alpes (ANR-19-P3IA-0003).

\bibliography{document_nmt}
\bibliographystyle{acl_natbib}


\appendix


\section{Context-discounted loss}\label{app:se:cd}

In CANMT with sliding concatenation windows we should prioritize the quality of the translation of the current sentence because the context translation will be discarded during inference.
Therefore, the standard NMT objective function is not suitable in this case.
\citet{lupo_focused_2022} propose to encourage the concatenation approach to focus on the translation of the current sentence $\bm{x}^j$ by applying a discount $0\leq{\small\text{CD}}<1$ to the loss generated by context tokens:

\begin{align} \label{eq:cd}
	\mathcal{L}_{\small\text{CD}}(\bm{x}_K^j,\bm{y}_K^j)
	&={\small\text{CD}\cdot}\mathcal{L}_{context} + \mathcal{L}_{current}\\
	&={\small\text{CD}\cdot}\mathcal{L}(\bm{x}_K^j,\bm{y}_{K-1}^{j-1}) + \mathcal{L}(\bm{x}_K^j,\bm{y}^j). \nonumber
\end{align}

with $\mathcal{L}(\bm{x},\bm{y})$ being the standard NMT objective function:

\begin{align}\label{eq:concatlikelihood}
	\mathcal{L}(\bm{x},\bm{y})
	&= \sum_{t=1}^{|\bm{y}|} \log P(y_t|\bm{y}_{<t},\bm{x}),
\end{align}

The authors demonstrate the efficacy of this loss function, that leads to a self-attentive mechanism that is less influenced by noisy contextual information.
As a result, they show a marked improvement in the translation of inter-sentential discourse phenomena.

\section{Details on experimental setup}\label{app:se:setup}

\begin{table*}[]
	\centering
	\begin{adjustbox}{width=\textwidth}
		\begin{tabular}{ccccccccccccc}
			\toprule
			Corpus & Tgt & Docs & Sents & \multicolumn{3}{c}{Doc Length} & \multicolumn{3}{c}{Sent Length} & \multicolumn{3}{c}{Sent Length (BPE)} \\
			\cmidrule(lr){5-7}
			\cmidrule(lr){8-10}
			\cmidrule(lr){11-13}
			&  &  &  & mean & std & max & mean & std & max & mean & std & max \\
			\midrule
			Voita & Ru & 1.5M & 6.0M & 4.0 & 0.0 & 4 & 8.3 & 4.7 & 64 & 8.6 & 4.9 & 69 \\
			IWSLT17 & De & 1.7k & 0.2M & 117.0 & 58.4 & 386 & 20.8 & 14.3 & 153 & 23.3 & 16.3 & 195 \\
			High & De & 12.2k & 2.3M & 188.4 & 36.2 & 386 & 27.3 & 16.1 & 249 & 29.1 & 17.4 & 408 \\
			\midrule
			Voita & Ru & 10k & 40k & 4.0 & 0.0 & 4 & 8.2 & 4.8 & 50 & 8.5 & 5.0 & 58 \\
			Both & De & 62 & 5.4k & 87.6 & 53.5 & 296 & 19.0 & 12.5 & 114 & 21.1 & 14.0 & 132 \\
			\midrule
			Voita & Ru & 10k & 40k & 4.0 & 0.0 & 4 & 8.2 & 4.8 & 42 & 8.5 & 5.0 & 50 \\
			Both & De & 12 & 1.1k & 90.0 & 29.2 & 151 & 19.3 & 12.7 & 102 & 21.6 & 14.3 & 116 \\
			\bottomrule
		\end{tabular}
	\end{adjustbox}
	\caption{Statistics for the training (1st block), validation (2nd block) and test set (3rd block) after pre-processing, and after BPE tokenization. All figures refer to the English text (source side).}
	\label{tab:se:datastats}
\end{table*}

All experiments are implemented in \textit{fairseq}~\cite{ott_fairseq_2019}. All models follow the \textit{Transformer-base} architecture \cite{vaswani_attention_2017}: hidden size of 512, feed forward size of 2048,  6 layers, 8  attention heads.
They are trained on 4 Tesla V100, with a fixed batch size of approximately 32k tokens for En$\rightarrow$Ru and 16k for En$\rightarrow$De, as it has been shown that Transformers need a large batch size to optimize performance \citep{popel_training_2018}.
We stop training after 12 consecutive non-improving validation steps (in terms of loss on dev), and we average the weights of the best-performing checkpoint and the 4 checkpoints that follow it. We train models with the optimizer configuration and learning rate (LR)  schedule described in \citet{vaswani_attention_2017}. The maximum LR is optimized for each model over the search space $\{7e-4,9e-4,1e-3,3e-3\}$.
The LR achieving the best loss on the validation set after convergence was selected. We use label smoothing with an epsilon value of 0.1~\citep{Pereyra_regularizing_2017} for all settings.
We adopt strong model regularization (dropout=0.3) following~\citet{kim_when_2019} and \citet{ma_comparison_2021}.
At inference time, we use beam search with a beam of 4 for all models. We adopt a length penalty of 0.6 for all models.
The other hyperparameters were set according to the relevant literature~\citep{vaswani_attention_2017, popel_training_2018, voita_when_2019, ma_comparison_2021, lopes_document-level_2020}.
When experimenting with segment-shifted position embeddings, the shift is equal to the average sentence length calculated over the training data, following \cite{lupo_focused_2022}.
In particular, we set shift$=8$ for En$\rightarrow$Ru, shift$=21$ for En$\rightarrow$De.

\subsection{Data pre-processing}

Since Voita's data have already been pre-processed \cite{voita_when_2019}, we only apply byte pair encoding \cite{sennrich_neural_2016} with 32k merge operations jointly for English and Russian.
For IWSLT17, instead, we tokenize data with the Moses toolkit~\cite{koehn_moses_2007}, clean them by removing long sentences, and encode them with byte pair encoding. The byte pair encoding is learned on the En$\rightarrow$De training data released by WMT17 for the news translation task using 32k merge operations jointly for source and target languages, to be compatible with the experiments presented in the next section of the Appendix (\ref{app:se:nei}).

\section{Increasing training data for the English to German pair}\label{app:se:nei}

\begin{table}[]
	\centering
	\begin{adjustbox}{width=\linewidth}
		\begin{tabular}{lccccc}
			\toprule
			System & Enc. & Pers. & PSE & CP & BLEU \\
			\midrule
			s4to4+\texttt{CD} &  &  &  & \textbf{82.24} & 31.69 \\
			s4to4+\texttt{CD} & shift & \checkmark &  & 80.45 & 30.71 \\
			s4to4+\texttt{CD} & sin & \checkmark & \checkmark & 80.85 & 31.40 \\
			s4to4+\texttt{CD} & lrn & \checkmark &  & 79.82 & 31.58 \\
			\bottomrule
		\end{tabular}
	\end{adjustbox}
	\caption{Context-discounted s4to4 trained on the En$\rightarrow$De high-resource setting, evaluated with the accuracy on ContraPro (CP) and BLEU on the test set.}\label{tab:se:nei}
\end{table}

We hypothesize that the model does not undergo sufficient training in the En$\rightarrow$De setting to reap the benefits of segment embeddings. Indeed, the training data volume is smaller than in the En$\rightarrow$Ru setting: 0.2 million sentences versus 6 million (see Table~\ref{tab:se:datastats}).
Therefore, we choose to experiment with more En$\rightarrow$De training data, employing the same high-resource setting of \citet{lupo_divide_2022}.
This setting expands the IWSLT17 training data \cite{cettolo_wit3_2012} by adding the News-Commentary-v12 and Europarl-v7 sets released by WMT17\footnote{http://www.statmt.org/wmt17/translation-task.html}.
The resulting training set comprises 2.3M sentences (see statistics in Table~\ref{tab:se:datastats}).
Training on this data is more expensive than training on the En$\rightarrow$Ru setting, considering that the average sentence length is 27.3 tokens versus 8.3 tokens, respectively.
Therefore, we only train the most promising approaches.\footnote{We set shift$= 27$ for segment-shifted position embeddings, consistently with the average sentence length of the training data.}
Their performances are compared in Table~\ref{tab:se:nei}.
As expected, the s4to4+CD model drastically improves its performance compared to training on IWSLT17 alone: +7.93 accuracy points on ContraPro and +2.37 BLEU points on the test set (c.f. Table~\ref{tab:se:cd_de}).
However, even with larger training volumes, segment position encodings do not seem to help s4to4+\texttt{CD} on the En$\rightarrow$De language pair.

\section{Allocating more space to segments in PSE}\label{se:exp:128}

For the En$\rightarrow$Ru language pair, we have found that one-hot and sinusoidal segment embeddings need to be integrated into PSE for being leveraged by s4to4+CD (Section~\ref{se:exp:main:cd}).
Instead, learned embeddings worked best when added to position embeddings. \\
Here, we evaluate whether PSE with learned segment embeddings would perform better if more dimensions were allocated to segments. In particular, we let the model learn to represent sentence positions in $d_{SE}=128$ dimensions, which leaves $d_{PE} = d_{model} - d_{SE} = 384$ dimensions to position embeddings, largely enough as shown in Section~\ref{se:approaches:pse}. \\
As shown in Table~\ref{tab:se:128}, increasing the number of dimensions allocated to segment embeddings deteriorates the performance on Voita's contrastive set.
The reason could simply be that adding more learnable parameters makes the task harder.

\begin{table*}
	\centering\scalebox{1}{
		\begin{tabular}{lccccccc:cc}
			\toprule
			System & Enc. & Pers. & PSE & Deixis & Lex co. & Ell. inf & Ell. vp & Voita & BLEU \\
			\midrule
			s4to4+\texttt{CD} & lrn & \checkmark & 4 & 93.20 & 47.40 & 72.20 & 64.40 & 74.48 & 32.35 \\
			s4to4+\texttt{CD} & lrn &  & 128 & 83.88 & 46.33 & 65.20 & 50.20 & 67.38 & 32.43 \\
			s4to4+\texttt{CD} & lrn & \checkmark & 128 & 78.20 & 46.40 & 40.60 & 30.60 & 60.14 & 32.35 \\
			\bottomrule
		\end{tabular}
	}\caption{s4to4 trained on En$\rightarrow$Ru OpenSubtitles. Accuracy on Voita's En$\rightarrow$Ru contrastive set and BLEU on the test set. The accuracy on the contrastive set is detailed on the left, with the accuracy on each subset corresponding to a specific discourse phenomenon. Result: allocating more dimensions to segments in PSE deteriorates performance.}\label{tab:se:128}
\end{table*}

\section{Persistent positions}\label{app:se:perspos}

\begin{table}[h]
	\centering
	\begin{adjustbox}{width=\linewidth}
		\begin{tabular}{lccccc}
			\toprule
			System & Enc. & Pers. & PSE & Voita & BLEU \\
			\midrule
			s4to4 &  &  &  & 72.02 & 32.45 \\
			s4to4 &  & \checkmark &  & \textbf{72.44} & 32.29 \\
			\midrule
			s4to4+\texttt{CD} &  &  &  & 73.42 & 32.37 \\
			s4to4+\texttt{CD} &  & \checkmark &  & 74.10 & 32.12 \\
			s4to4+\texttt{CD} & shift & \checkmark &  & \textbf{75.94} & 31.98 \\
			\bottomrule
		\end{tabular}
	\end{adjustbox}
	\caption{En$\rightarrow$Ru: making positions persistent across Transformer's blocks improve discourse disambiguation performance both for vanilla and context-discounted s4to4. Segment-shifting positions further improves performance.}\label{tab:se:perspos}
\end{table}

Making sentence position encodings persistent across the layers have been found beneficial for context-discounted models on the En$\rightarrow$Ru setting (Table~\ref{tab:se:cd_ru}).
The best-performing model, s4to4+CD+shift+pers, shifts token positions by a constant factor every time we pass from one sentence to the next and makes the resulting position embeddings persistent throughout Transformer's blocks.
In Table~\ref{tab:se:perspos}, we benchmark this model against models employing persistent token position embeddings but without segment-shifting. Both vanilla and context-discounted s4to4 perform better when positions are persistent across Transformer's blocks, as suggested by \citet{liu_learning_2020} and \citet{chen_breaking_2021}.
Segment-shifting further enhances performance, which confirms that the model benefits from a sharper distinction between sentences.

\section{Details of the evaluation on discourse phenomena}\label{app:se:discourse}

In Tables \ref{tab:se:discdetails_ru} and \ref{tab:se:discdetails_de}, we provide more details on the evaluation of the models presented in the tables of the paper, documenting their accuracy on the different subsets of the contrastive sets employed.
For Voita's En$\rightarrow$Ru contrastive set \cite{voita_when_2019}, we report the accuracy on each of the 4 discourse phenomena included in it; for the En$\rightarrow$De ContraPro (CP, \citet{muller_large-scale_2018}), the accuracy on anaphoric pronouns with antecedents at different distances $d=1,2,...$ (in number of sentences).
We complement Voita/CP with two other metrics, Voita/CP$_{\text{avg}}$ and CP$_{d>0}$. Metrics are calculated as follow:

\begin{equation}
	\begin{adjustbox}{width=0.49\textwidth}
	$\text{Voita} = \frac{2500*\text{Deixis} + 1500*\text{Lex co.} + 500*\text{Ell. inf} + 500*\text{Ell. vp}}{5000}$
	\end{adjustbox}
\end{equation}

\begin{equation}
	\begin{adjustbox}{width=0.48\textwidth}
			$\text{CP}_{alld} = \frac{2400*\text{(d=0)} + 7075*\text{(d=1)} + 1510*\text{(d=2)} + 573*\text{(d=3)} + 442*\text{(d>3)}}{12000}$
	\end{adjustbox}
\end{equation}

\begin{equation}
	\begin{adjustbox}{width=0.48\textwidth}
		$\text{CP$_{d>0}$} = \frac{7075*\text{(d=1)} + 1510*\text{(d=2)} + 573*\text{(d=3)} + 442*\text{(d>3)}}{9600}$
	\end{adjustbox}
\end{equation}

\begin{equation}
	\small
	\text{Voita}_{\text{avg}}\text{/CP}_{\text{avg}} = \frac{\text{(d=1)} + \text{(d=2)} + \text{(d=3)} + \text{(d=4)}}{4}
\end{equation}


\begin{table*}
	\centering
	\begin{adjustbox}{width=\linewidth}
		\begin{tabular}{lccccccc:cc}
			\toprule
			System & Enc. & Pers. & PSE & Deixis & Lex co. & Ell. inf & Ell. vp & Voita & Voita$_{\text{avg}}$\\
			\midrule
			base &  &  &  & 50.00 & 45.87 & 51.80 & 27.00 & 46.64 & 43.67 \\
			s4to4 & \textbf{} & \textbf{} & \textbf{} & 85.80 & 46.13 & 79.60 & 73.20 & 72.02 & 71.18 \\
			\midrule
			s4to4 & shift &  &  & 85.24 & 46.07 & 77.20 & 71.20 & 71.28 & 69.93 \\
			s4to4 & shift & \checkmark &  & \textbf{85.96} & \textbf{46.33} & 75.20 & \textbf{74.00} & 71.80 & 70.37 \\
			\midrule
			s4to4 & sin &  &  & \textbf{86.36} & 45.80 & 76.40 & 73.60 & 71.92 & 70.54 \\
			s4to4 & sin & \checkmark &  & 84.96 & 46.13 & 74.80 & 74.00 & 71.20 & 69.97 \\
			s4to4 & sin &  & \checkmark & 84.64 & \textbf{46.40} & 76.60 & 73.60 & 71.26 & 70.31 \\
			s4to4 & sin & \checkmark & \checkmark & 85.24 & 46.33 & 76.40 & \textbf{75.20} & 71.68 & 70.79 \\
			\midrule
			s4to4 & lrn &  &  & 85.48 & 46.27 & 76.20 & \textbf{75.60} & 71.80 & 70.89 \\
			s4to4 & lrn & \checkmark &  & 84.84 & 45.93 & 77.60 & 74.40 & 71.40 & 70.69 \\
			s4to4 & lrn &  & \checkmark & 83.60 & \textbf{46.67} & 74.80 & 70.80 & 70.36 & 68.97 \\
			s4to4 & lrn & \checkmark & \checkmark & \textbf{90.52} & 46.00 & 74.80 & 66.60 & \textbf{73.20} & 69.48 \\
			\midrule
			s4to4 & 1hot &  &  & \textbf{86.08} & 47.07 & 78.00 & \textbf{75.60} & \textbf{72.52} & \textbf{71.69} \\
			s4to4 & 1hot & \checkmark &  & 83.76 & \textbf{47.53} & 78.00 & 75.00 & 71.44 & 71.07 \\
			s4to4 & 1hot &  & \checkmark & 84.56 & 46.13 & 78.20 & 73.00 & 71.24 & 70.47 \\
			s4to4 & 1hot & \checkmark & \checkmark & 84.56 & 46.47 & 76.00 & 73.40 & 71.16 & 70.11 \\
			\midrule
			s4to4+\texttt{CD} &  &  &  & 87.16 & 46.40 & 81.00 & 78.20 & 73.42 & 73.19 \\
			\midrule
			s4to4+\texttt{CD} & shift &  &  & 85.76 & 48.33 & 81.40 & 80.40 & 73.56 & 73.97\\
			s4to4+\texttt{CD} & shift & \checkmark &  & \textbf{88.76} & \textbf{52.13} & \textbf{83.00} & 76.20 & \textbf{75.94} & \textbf{75.02} \\
			\midrule
			s4to4+\texttt{CD} & sin &  &  & 87.96 & 46.80 & 78.00 & 76.60 & 73.48 & 72.34 \\
			s4to4+\texttt{CD} & sin & \checkmark &  & 86.80 & \textbf{47.00} & 80.80 & 78.20 & 73.40 & 73.20 \\
			s4to4+\texttt{CD} & sin & \checkmark & \checkmark & \textbf{89.28} & 46.67 & \textbf{83.20} & 77.20 & \textbf{74.68} & \textbf{74.09} \\
			\midrule
			s4to4+\texttt{CD} & lrn &  &  & 88.12 & 46.47 & 81.20 & 75.60 & 73.68 & 72.85 \\
			s4to4+\texttt{CD} & lrn & \checkmark &  & 86.84 & \textbf{52.27} & \textbf{84.60} & \textbf{80.00} & \textbf{75.56} & \textbf{75.93} \\
			s4to4+\texttt{CD} & lrn & \checkmark & \checkmark & \textbf{93.20} & 47.40 & 72.20 & 64.40 & 74.48 & 69.30\\
			\midrule
			s4to4+\texttt{CD} & 1hot &  &  & 86.40 & 46.73 & 82.00 & 76.40 & 73.06 & 72.88 \\
			s4to4+\texttt{CD} & 1hot & \checkmark &  & 87.68 & 46.80 & 81.60 & \textbf{78.60} & 73.90 & 73.67\\
			s4to4+\texttt{CD} & 1hot & \checkmark & \checkmark & \textbf{88.88} & \textbf{47.67} & \textbf{82.20} & 75.40 & \textbf{74.50} & 73.54\\
			\midrule
			Sample size & & & & 2500 & 1500 & 500 & 500 & 5000 & 5000 \\
			\bottomrule
		\end{tabular}
	\end{adjustbox}
	\caption{Accuracy on the En$\rightarrow$Ru contrastive set for the evaluation of discourse phenomena (Voita, \%), and on its 4 subsets: deixis, lexical cohesion, inflection ellipsis, and verb phrase ellipsis. Voita$_{\text{avg}}$ denotes the average on the 4 discourse phenomena, while Voita represents the average weighted by the frequency of each phenomenon in the test set (see row "Sample size"). }\label{tab:se:discdetails_ru}
\end{table*}

\begin{table*}
	\centering
	\begin{adjustbox}{width=\linewidth}
		\begin{tabular}{lcccccccc:ccc}
			\toprule
			System & Enc. & Pers. & PSE & d=0 & d=1 & d=2 & d=3 & d>3 & CP$_{d>0}$ & CP$_{\text{avg}}$ & CP \\
			\midrule
			base &  &  &  & 68.75 & 32.89 & 43.97 & 47.99 & 70.58 & 37.27 & 48.86 & 43.57 \\
			s4to4 &  &  &  & 75.20 & 68.89 & 74.96 & 79.58 & 87.78 & 71.35 & 77.80 & 72.12 \\
			s4to4+\texttt{CD} &  &  &  & 76.66 & 72.86 & 75.96 & 80.10 & 84.38 & 74.31 & 78.33 & 74.78 \\
			\midrule
			s4to4+\texttt{CD} & shift &  &  & 75.25 & 72.56 & 77.15 & 80.27 & 86.65 & 74.39 & 79.16 & 74.56 \\
			s4to4+\texttt{CD} & shift & \checkmark &  & 72.41 & 69.15 & 74.23 & 77.13 & 86.42 & 71.22 & 76.73 & 71.46 \\
			\midrule
			s4to4+\texttt{CD} & sin &  &  & 76.75 & 71.83 & 76.82 & 80.97 & 87.55 & 73.88 & 79.29 & 74.46 \\
			s4to4+\texttt{CD} & sin & \checkmark &  & 76.50 & 72.08 & 76.35 & 79.23 & 85.97 & 73.82 & 78.41 & 74.35 \\
			s4to4+\texttt{CD} & sin & \checkmark & \checkmark & 77.25 & 71.22 & 76.42 & 78.88 & 86.87 & 73.22 & 78.35 & 74.02 \\
			\midrule
			s4to4+\texttt{CD} & lrn &  &  & 73.91 & 70.21 & 75.29 & 77.66 & 85.06 & 72.14 & 77.06 & 72.49 \\
			s4to4+\texttt{CD} & lrn & \checkmark &  & 73.66 & 68.53 & 72.51 & 75.74 & 86.65 & 70.42 & 75.86 & 71.07 \\
			s4to4+\texttt{CD} & lrn & \checkmark & \checkmark & 73.54 & 68.40 & 79.07 & 80.27 & 83.48 & 71.48 & 77.81 & 71.89 \\
			\midrule
			\multicolumn{12}{c}{High Resource Setting} \\
			\midrule
			base &  &  &  & 82.83 & 35.18 & 44.90 & 51.13 & 66.28 & 39.09 & 49.37 & 47.84 \\
			s4to4 &  &  &  & 82.41 & 80.66 & 81.72 & 84.29 & 88.00 & 81.38 & 83.67 & 81.59 \\
			s4to4+\texttt{CD} &  &  &  & 83.70 & 81.79 & 82.11 & 82.19 & 90.04 & 82.24 & 84.03 & 82.54 \\
			\midrule
			s4to4+\texttt{CD} & shift & \checkmark &  & 81.70 & 79.61 & 81.45 & 83.42 & 86.65 & 80.45 & 82.78 & 80.70 \\
			s4to4+\texttt{CD} & sin & \checkmark & \checkmark & 84.12 & 79.85 & 82.38 & 84.46 & 86.87 & 80.85 & 83.39 & 81.50 \\
			s4to4+\texttt{CD} & lrn & \checkmark &  & 83.12 & 79.13 & 79.73 & 82.19 & 88.00 & 79.82 & 82.26 & 80.48 \\
			\midrule
			Sample size & & & & 2400 & 7075 & 1510 & 573 & 442 & 9600 & 9600 & 12000 \\
			\bottomrule
		\end{tabular}
	\end{adjustbox}
	\caption{Accuracy on the En$\rightarrow$De contrastive set for the evaluation of anaphoric pronouns (CP = ContraPro, \%). The columns titled d=* represent the accuracy for each subset of pronouns with antecedents at a specific distance $d\in[0,1,2,3,>3]$ (in number of sentences). CP$_{\text{avg}}$ denotes the average on the 4 subsets of pronouns with extra-sentential antecedents ($d>0$) while CP$_{d>0}$ represents the average weighted by the size of each of the 4 subsets (see row "Sample size"). CP is equivalent to CP$_{d>0}$, but it includes the accuracy on $d=0$.}\label{tab:se:discdetails_de}
\end{table*}

\end{document}